\documentclass[journal]{IEEEtran}
\usepackage{xcolor,soul,framed} 

\colorlet{shadecolor}{yellow}
\usepackage[pdftex]{graphicx}
\graphicspath{{../pdf/}{../jpeg/}}
\DeclareGraphicsExtensions{.pdf,.jpeg,.png}

\usepackage[cmex10]{amsmath}
\usepackage{array}
\usepackage{mdwmath}
\usepackage{mdwtab}
\usepackage{eqparbox}
\usepackage{url}

\usepackage{amssymb}
\usepackage{multirow}
\newcommand{\RNum}[1]{\uppercase\expandafter{\romannumeral #1\relax}}
\usepackage[ruled]{algorithm2e}
\usepackage{hyperref}
\usepackage{CJKutf8}
\usepackage{makecell}
\hypersetup{
    colorlinks=true,            
    linkcolor=blue,             
    filecolor=cyan,             
    urlcolor=magenta,           
    pdftitle={Overleaf Example},
    pdfpagemode=FullScreen,
    }
    
\begin{document}
\bstctlcite{IEEEexample:BSTcontrol}
    \title{A Chinese Continuous Sign Language Dataset Based on Complex Environments}
  \author{
   \IEEEauthorblockN{Jing~Li$^{1}$,
    Qidan~Zhu$^{2}$,
    Linglian~Li$^{3*}$,
    Fei~Yuan$^{2}$, 
    Quan~Gan$^2$}

    \IEEEauthorblockA{$^1$ School of Automation,Chengdu University of Information Technology,Chengdu 610225,China}\par
     \IEEEauthorblockA{$^2$ College of Intelligent Systems Science and Engineering, Harbin Engineering University, Harbin, 150001, China}\par
      \IEEEauthorblockA{$^3$ The Fifth Electronics Research Institute of Ministry of Industry and Information Technology, GuangZhou, 511370, China}\par

 \thanks{*Corresponding author\par
Email addresses: \par
ljing@hrbeu.edu.cn (Jing Li), \par
zhuqidan@hrbeu.edu.cn (Qidan Zhu), \par
13922420121@163.com (Linglian Li), \par
bohelion@hrbeu.edu.cn (Fei Yuan), \par
gquan@hrbeu.edu.cn (Quan Gan)}
  }

\maketitle

\begin{abstract}
The current bottleneck in continuous sign language recognition (CSLR) research lies in the fact that most publicly available datasets are limited to laboratory environments or television program recordings, resulting in a single background environment with uniform lighting, which significantly deviates from the diversity and complexity found in real-life scenarios. To address this challenge, we have constructed a new, large-scale dataset for Chinese continuous sign language (CSL) based on complex environments, termed the complex environment - Chinese sign language dataset (CE-CNSL). This dataset encompasses 5,988 continuous CSL video clips collected from daily life scenes, featuring more than 70 different complex backgrounds to ensure representativeness and generalization capability. We also propose a time-frequency sequence based continuous sign language recognition model (TFNet) to address the issue of advanced CSLR methods ignoring frequency domain features. After frame level feature extraction, this model utilizes both time-domain and frequency-domain information to extract sequence features separately, and then fuses them to achieve efficient and accurate continuous sign language recognition. Experimental results demonstrate that our approach achieves significant performance improvements on the CE-CNSL, validating its effectiveness under complex background conditions. Additionally, our proposed method has also yielded highly competitive results when applied to three publicly available CSL datasets. Code and dataset is available at \url{https://github.com/woshisad159/TFNet.git}.
\end{abstract}


%
\IEEEpeerreviewmaketitle


\section{Introduction}

\IEEEPARstart{S}{i}gn language recognition has always been a key issue of concern for researchers. As a model of interdisciplinary research, it integrates the essence of computer vision, natural language processing, and human-computer interaction technology\cite{pu2020boosting}\cite{wadhawan2021sign}. Sign language recognition technology not only simplifies the complexity of cross linguistic communication by translating sign language dynamic videos into understandable text or speech in real time, but also demonstrates enormous potential in multiple fields such as education, employment, and public services\cite{cui2017recurrent}.\par

In the field of CSLR, achieving accurate recognition of continuous sign language relies heavily on having a high-quality CSL dataset\cite{uthus2024youtube}\cite{duarte2021how2sign}\cite{ozdemir2020bosphorussign22k}\cite{wang2014similarity}. There are currently four commonly used publicly available sign language recognition datasets. Koller et al.\cite{koller2015continuous} constructed the PHOENIX14-2014 dataset by collecting weather forecast programs broadcasted by a German public television station, providing valuable resources for CSLR research. This dataset is based on demonstrations by professional sign language interpreters; however, its collection environment remains limited to the television studio, lacking the diversity of real-world contexts. Camgoz et al.\cite{camgoz2018neural} further expanded upon this by introducing the PHOENIX14-2014T dataset, increasing the volume of data and refining the definition of sign boundaries, thereby enhancing annotation accuracy. Although this dataset improves in terms of professionalism, it is still constrained by relatively idealized background conditions, making it difficult to comprehensively reflect the sign language communication scenarios of everyday life. Adaloglou et al.\cite{adaloglou2021comprehensive}, focusing on practical application scenarios, particularly emphasized communication between deaf individuals and public service institutions, such as police departments, hospitals, and citizen service centers. They meticulously designed five typical scenarios, simulated dialogues of varying difficulty levels, thus enhancing the practicality and authenticity of the dataset. Zhou et al.\cite{zhou2021improving} constructed a large-scale, continuous dataset of Chinese sign language, focusing on people's daily life scenarios. This dataset is based on demonstrations by professional sign language interpreters and covers multiple topics such as family life, healthcare, and school life. Although the above-mentioned datasets have played an important role in advancing continuous sign language recognition technology, they generally face a common challenge: the collection environment is disconnected from real life, making it difficult to fully reflect sign language communication scenarios in daily life, as shown in Figure 1.\par

\begin{figure*}
  \begin{center}
  \includegraphics[width=3.5in]{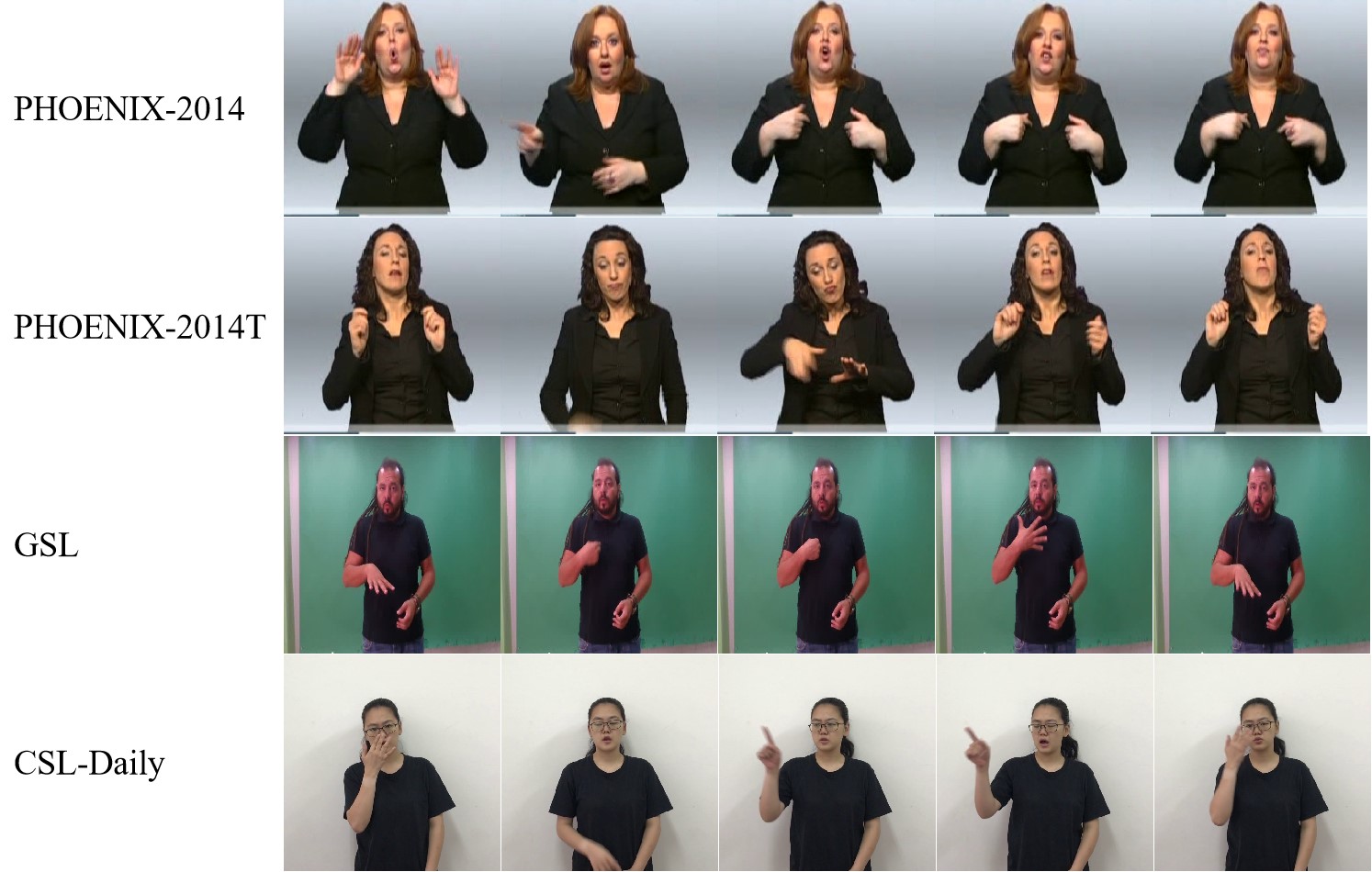}\\
  \caption{The above are partial image sequences from four publicly available continuous sign language datasets, PHOENIX14-2014, PHOENIX14-2014T, GSL, and CSL-Daily. It can be seen from these image sequences that these datasets were collected in laboratory environments or television programs, with the background of the subjects being a solid color background and uniform lighting}\label{fig:ljxy1}
  \end{center}
\end{figure*}

To overcome the limitations of existing continuous sign language datasets, particularly their disconnection from real-life scenarios, this study constructs a CSL dataset oriented towards practical application environments. The dataset aims to facilitate the seamless transition of CSLR technology from laboratory settings to everyday life, laying a solid foundation for barrier-free communication between the deaf and hearing communities. The construction process of the CE-CNSL dataset strictly adheres to the principle of practical application orientation, collecting a large number of continuous sign language video materials derived from real-life scenarios, covering a wide range of situational variations and environmental complexities. After the completion of the CE-CNSL dataset, we conducted a comprehensive evaluation using advanced deep learning models to verify its reliability and value as a research benchmark.\par

\begin{figure*}
  \begin{center}
  \includegraphics[width=5.5in]{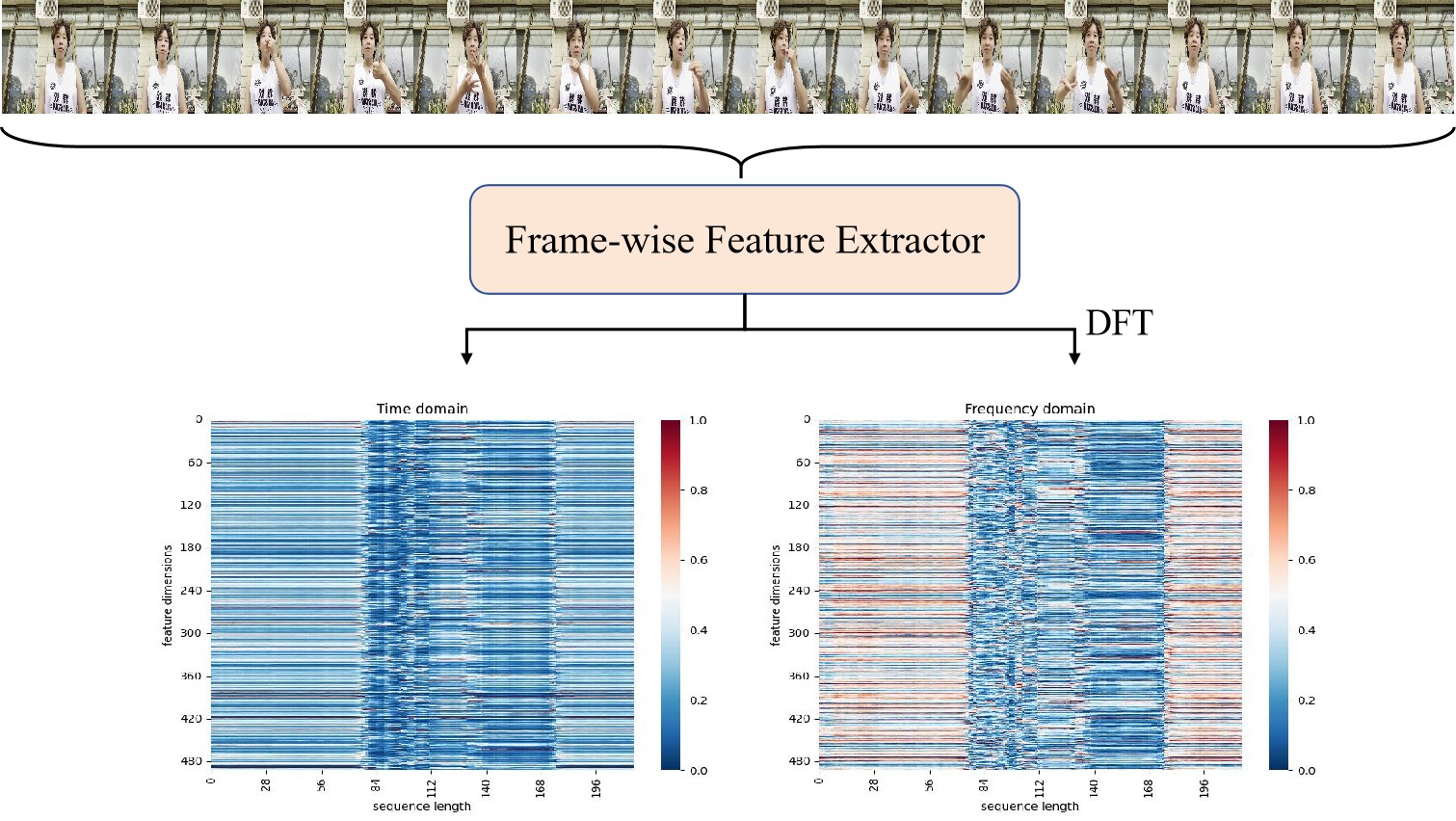}\\
  \caption{Heat maps of feature vector sequences in both time and frequency domains for a sign language video after frame level feature extraction}\label{fig:ljxy2}
  \end{center}
\end{figure*}

In addition, temporal feature extraction is a crucial step in CSLR tasks. Sign language, as a highly dynamic visual language, relies not only on the static shape of gestures, but also on the motion trajectory, temporal information, and dynamic features of gestures. These dynamic features change over time and can describe the evolution of sign language expression. However, the current method \cite{xie2023multi}\cite{wei2020semantic}\cite{aditya2022novel} only considers the temporal aspect and ignores the characteristics in the frequency domain. The frequency domain based sign language features can provide another perspective on the periodic motion patterns and rhythm changes in sign language. Different feature forms can greatly enrich the expression of sign language features, as shown in Figure 2.\par

This paper proposes a new sequence feature extraction module based on the consideration of feature diversity. By using 1DCNN BILSTM for sequence feature modeling in both time and frequency domains, rich sign language feature expressions are obtained by integrating information from different domains. An end-to-end continuous sign language recognition model based on time-frequency sequences (TFNet) is also proposed. This combination of cross domain features not only enriches the model's feature information, but also enhances its ability to recognize complex patterns, thereby demonstrating higher performance in various application scenarios. The experimental results demonstrate that TFNet not only achieves highly competitive results on currently publicly available datasets, but also shows significant performance improvements compared to existing advanced models on the CE-CNSL dataset.\par

In summary, the main contributions of this paper are as follows:\par

\begin{itemize}
\item[$\bullet$] We have constructed a Chinese Continuous Sign Language dataset CE-CNSL dataset oriented towards practical application environments. This dataset, with its rich content and precise annotations, aims to provide a robust foundational resource for the advancement of CSLR technology.
\item[$\bullet$] We propose a new sequence feature extraction module that innovatively extracts frequency domain features. By integrating different domain feature information, enriching feature expression, and obtaining efficient and accurate semantic expression, it provides a new perspective for exploring the periodic motion patterns and rhythm changes in sign language. It can effectively reveal the imperceptible periodic patterns in sign language actions and significantly improve the accuracy of continuous sign language recognition in complex environments.
\item[$\bullet$] We propose a continuous sign language recognition model based on time-frequency sequences (TFNet), which greatly expands the dimensions of comprehensive understanding of sign language features, providing indispensable support for a deeper understanding of the prosodic structure of sign language and improving the accuracy of sign language recognition. It has high universality and robustness on different continuous sign language recognition datasets.
\end{itemize}

\begin{table*}[!htbp]
\centering
\caption{Large-scale publicly available continuous sign language datasets based on video.}
\label{tab:aStrangeTable1}
\begin{tabular}{c|c|c|c|c|c|c|c|c|c}
\hline  
Dataset & Language & Signers & Videos & Duration (hours) & Vocab & Resolution & Modalities & FPS & Source \\
\hline
SIGNUM\cite{von2010signum} & German & 25 & 11,375 & 8.43 & 455 & $776 \times 578$ & RGB & 30 & Lab \\
\hline  
PHOENIX-2014\cite{koller2015continuous} & German & 9 & 6,841 & 10.71 & 1,231 & $210 \times 260$ & RGB & 25 & TV \\
\hline 
PHOENIX-2014T\cite{camgoz2018neural} & German & 9 & 8,257 & 10.53 & 1,231 & $210 \times 260$ & RGB & 25 & TV \\
\hline 
GSL\cite{adaloglou2021comprehensive} & Greek & 7 & 10,295 & 9.59 & 310 & $848 \times 480$ & RGB & 30 & Lab \\
\hline 
CSL\cite{huang2018video} & Chinese & 50 & 25,000 & 100+ & 178 & $1280 \times 720$ & RGB+D & 30 & Lab \\
\hline 
CSL-Daily\cite{zhou2021improving} & Chinese & 10 & 20,654 & 23.27 & 2,000 & $1920 \times 1080$ & RGB+D & 30 & Lab \\
\hline 
TVB-HKSL-News\cite{niu2024hong} & Chinese & 2 & 7,160 & 16.07 & 6,515 & $248 \times 360$ & RGB & 25 & TV \\
\hline 
CE-CNSL(ours) & Chinese & 12 & 5,988 & 10.52 & 3,515 & varying & RGB & varying & Lives \\
\hline  
\end{tabular}
\end{table*}

\section{Related Work}

\subsection {Continuous sign language dataset}

A CSL dataset serves as an indispensable cornerstone for researching CSLR technology, with its quality directly determining the performance and application potential of recognition algorithms\cite{zhou2021improving}\cite{feng2022sign}. In recent years, with the rapid development of artificial intelligence technologies, a series of CSL datasets have emerged, contributing significantly to the advancement of this field. Agris et al.\cite{von2010signum} pioneered the creation of the SIGNUM sign language dataset, which includes rich samples of isolated words and continuous sentences performed by 25 signers of different genders and age groups. Although SIGNUM made initial explorations in diversity and corpus size, the limitation of 450 different German sign language words and 780 continuous sentences did not adequately meet the large-scale data requirements of deep learning algorithms. In 2015, Koller et al.\cite{koller2015continuous} introduced the PHOENIX14-2014 dataset, which collected 6,841 continuous sentences through weather forecast programs on German public television, significantly enhancing the scale and practicality of the dataset. Building on this, Camgoz et al.\cite{camgoz2018neural} extended the dataset to create the PHOENIX14-2014T dataset, enhancing its professionalism and introducing refined sign boundary definitions, annotated by deaf experts, further improving the dataset's quality. Huang et al.\cite{huang2018video} focused on Chinese sign language, collecting video data including RGB, depth, and body joint patterns using Microsoft Kinect cameras, providing a multimodal perspective for sign language recognition research. Adaloglou et al.\cite{adaloglou2021comprehensive} concentrated on practical application scenarios of sign language, building cases involving interactions with police departments, hospitals, and citizen service centers, enriching the diversity of scenes in the dataset. Niu et al.\cite{niu2024hong} utilized an automated data collection pipeline to systematically gather extensive continuous sign language videos from seven months of sign language news broadcasts by Hong Kong Television Broadcasts Limited (TVB), further expanding the publicly available data resources. Despite the critical role these datasets have played in advancing sign language recognition technology, most originate from laboratory environments or television programs, commonly featuring a single background and uniform lighting conditions, which differ significantly from real-life sign language communication scenarios. In Table \RNum{1}, we list some large-scale publicly available continuous sign language datasets based on video. \par

Given the limitations of existing datasets, we have constructed a Chinese continuous sign language dataset that truly reflects the complexity of real life. Unlike being limited to laboratories or television programs in the past, our dataset is collected by sign language presenters in daily environments, aiming to capture sign language dynamics in real-life scenarios, including changing backgrounds, natural lighting conditions, and rich contextual changes.\par

\begin{table*}[!htbp]
\centering
\caption{Statistical Data of the CE-CNSL Dataset.(OOV: out-of-vocabulary, e.g., words that occur in Dev set or Test set but not in Train set.)}
\label{tab:aStrangeTable2}
\begin{tabular}{c|c|c|c}
\hline  
dataset split & train & dev & test \\
\hline 
signers & 12 & 12 & 12 \\
\hline 
duration[h] & 8.71 & 0.89 & 0.92 \\
\hline 
frames & 930,841 & 95,891 & 99,390 \\
\hline 
sentences & 4,973 & 515 & 500 \\
\hline 
average glosses & 5.69 & 5.52 & 6.08 \\
\hline 
vocabulary$\cdot$size & 3,515 & 774 & 714 \\
\hline 
total OOVs & - & 0 & 0 \\
\hline 
resolution & varying & varying & varying \\
\hline  
\end{tabular}
\end{table*}

\begin{table*}[!htbp]
\centering
\caption{Statistics of Recording Equipment, Age, and Gender for Sign Language Performers.}
\label{tab:aStrangeTable3}
\begin{tabular}{c|c|c|c|c|c|c}
\hline  
\multirow[t]{1}{*}{ signers } & \multicolumn{3}{c|}{ video numbers } & \multirow[t]{1}{*}{ phone types } & \multirow[t]{1}{*}{ ages } & \multirow[t]{1}{*}{ genders } \\
\cline{2-4}
& train & dev & test & & & \\
\hline 
A & 490 & 55 & 55 & iPhone 14 Plus & $20-25$ & $\operatorname{man}$ \\
\hline 
B & 490 & 53 & 57 & iPhone 12 Pro & $20-25$ & woman \\
\hline 
C & 504 & 54 & 42 & iPhone 13 & $20-25$ & man \\
\hline
D & 507 & 46 & 47 & iPhone 13 & $20-25$ & woman \\
\hline 
E & 495 & 51 & 54 & iPhone 13 & $20-25$ & $\operatorname{man}$ \\
\hline 
F & 492 & 50 & 58 & HUAWEI P50 Pro & $25-30$ & woman \\
\hline 
G & 490 & 54 & 56 & OnePlus Ace & $25-30$ & woman \\
\hline 
H & 101 & 9 & 9 & iPhone 13 Pro & $25-30$ & $\operatorname{man}$ \\
\hline 
I & 66 & 9 & 9 & iPhone 13 Pro & $40-45$ & woman \\
\hline 
J & 330 & 29 & 38 & iPhone 14 Pro Max & $20-25$ & woman \\
\hline 
K & 501 & 53 & 46 & iPhone 12 & $20-25$ & woman \\
\hline 
L & 507 & 52 & 29 & iPhone 14 Pro & $20-25$ & woman \\
\hline  
\end{tabular}
\end{table*}

\subsection {Continuous sign language recognition}

Effectively extracting and utilizing temporal features is of great significance for improving the accuracy and robustness of continuous sign language recognition\cite{rastgoo2021sign}\cite{cheng2020fully}\cite{hu2022temporal}. Hu et al.\cite{hu2023continuous} innovatively introduced a correlation module that can dynamically calculate the correlation mapping between the current frame and adjacent frames, effectively identifying the trajectory of spatial patches and providing a new perspective for continuous sign language recognition. Zhu et al.\cite{zhu2024continuous} focused on the dynamic attention mechanism, which can effectively capture the nonlinear changes in local motion areas during sign language expression, generate dynamic descriptions of image changes, and further enhance the model's ability to capture complex sign language dynamics. The above methods mainly focus on the motion trajectory between images, and improve the accuracy of sign language recognition by enhancing the ability to capture dynamic features. In terms of temporal feature extraction, only a hybrid convolutional neural network composed of 1DCNN and BiLSTM is used for feature extraction. Li et al.\cite{zhu2024multiscale} are committed to building a multi-scale temporal feature network, which significantly improves the flexibility and accuracy of the model in modeling sequences through different scales of temporal receptive fields. However, the introduction of multi-scale convolution kernels resulted in a sharp increase in the number of model parameters. Zhou et al.\cite{zhou2020spatial} proposed a Spatiotemporal Multi Clue (STMC) network to solve vision based sequence learning problems. The network consists of a Spatial Multi Clue (SMC) module and a Temporal Multi Clue (TMC) module. The introduction of multiple clues greatly improves the accuracy of the model, but it is also the use of multiple clues that increases the computational complexity and inference time of the model. These methods, like many existing methods\cite{yin2023gloss}\cite{koller2017re}\cite{hu2023signbert+}\cite{wei2023improving}, globally pool the extracted frame level features, convert them into one-dimensional sequence features related to time, and then perform sequence feature extraction. Only considering the temporal aspect, while ignoring the characteristics in the frequency domain. This paper proposes a continuous sign language recognition model based on time-frequency sequences (TFNet). This method aims to achieve efficient and accurate sign language recognition by jointly extracting sequence features in both time and frequency domains, opening up a new path for the development of continuous sign language recognition technology.\par

\begin{figure*}
  \begin{center}
  \includegraphics[width=6.0in,height=2.0in]{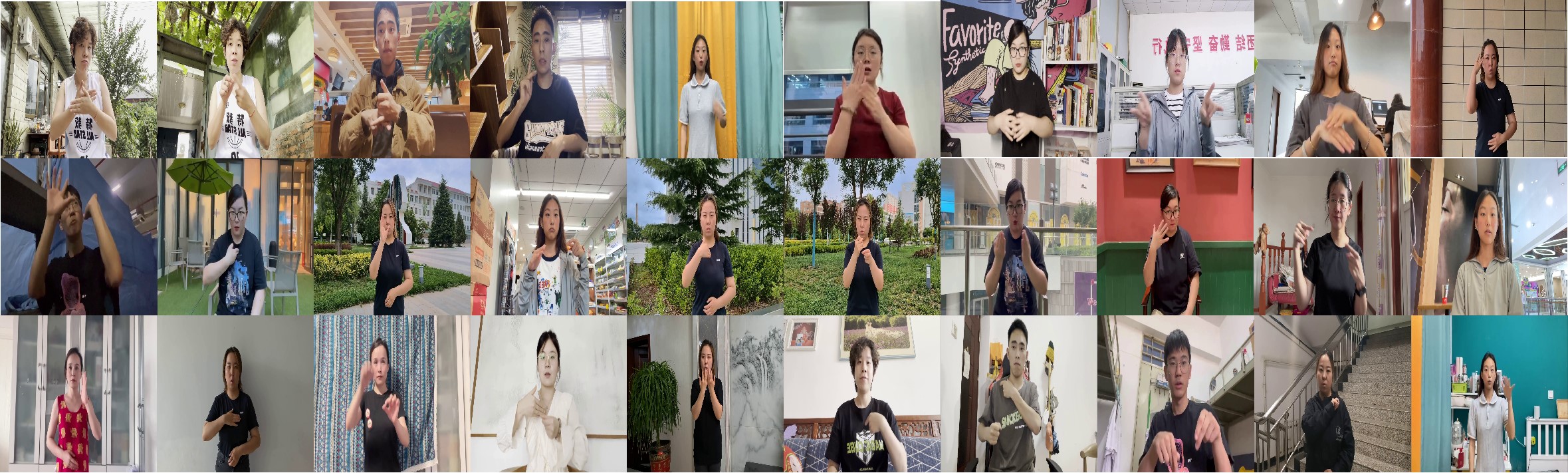}\\
  \caption{Illustration of backgrounds in selected images from the CE-CNSL dataset.}\label{fig:ljxy3}
  \end{center}
\end{figure*}

\section{Dataset and Methodology}

\subsection {CE-CNSL dataset}

To promote the transition of CSLR technology from laboratory environments to real-life applications, we have carefully constructed a publicly available Chinese Continuous Sign Language dataset named CE-CNSL. Unlike the limitations of previous datasets, CE-CNSL aims to simulate natural living scenarios, without restricting recording equipment, attire choices, or background environments, ensuring that the collected sign language videos closely resemble real-world application contexts.\par

\begin{table*}[!htbp]
\centering
\caption{Partial Annotations of Continuous Sign Language Videos. "Signers" indicates the signer's identifier, "Chinese Sentences" represents the unsegmented spoken language translation, "Gloss" denotes the segmented sign gloss, and "Note" specifies which words use regional sign language. Words highlighted in red are regional sign language terms.}
\label{tab:aStrangeTable4}
\begin{tabular}{c|c|c|c|c}
\hline  
Number & Signers & Chinese Sentences & Gloss & Note \\
\hline 
1 & A & \begin{CJK*}{UTF8}{gbsn}今晚瑜伽课什么时候开始呢?\end{CJK*} & \begin{CJK*}{UTF8}{gbsn}今天/晚上/\textcolor{red}{瑜伽}/课/开始/什么/时间/?\end{CJK*} & \begin{CJK*}{UTF8}{gbsn}\textcolor{red}{瑜伽}\end{CJK*} \\
\hline 
2 & A & \begin{CJK*}{UTF8}{gbsn}一大盆异常茂盛的法国吊兰。\end{CJK*} & \begin{CJK*}{UTF8}{gbsn}大/茂盛/法国/花/名字/\textcolor{red}{吊兰}/。\end{CJK*} & \begin{CJK*}{UTF8}{gbsn}\textcolor{red}{吊兰}\end{CJK*} \\
\hline 
3 & A & \begin{CJK*}{UTF8}{gbsn}人生有许多困难和失败。\end{CJK*} & \begin{CJK*}{UTF8}{gbsn}\textcolor{red}{人生}/困难/失败/多/。\end{CJK*} & \begin{CJK*}{UTF8}{gbsn}\textcolor{red}{人生}\end{CJK*} \\
\hline 
4 & A & \begin{CJK*}{UTF8}{gbsn}他送给我们一件他旅游的纪念品。\end{CJK*} & \begin{CJK*}{UTF8}{gbsn}他/旅游/纪念/\textcolor{red}{品}/送 (我)/。\end{CJK*} & \begin{CJK*}{UTF8}{gbsn}\textcolor{red}{品}\end{CJK*} \\
\hline  
\end{tabular}
\end{table*}

The CE-CNSL dataset encompasses 12 sign language performers (labeled A through L), including eight females and four males. Notably, performers I and J are hearing-impaired individuals, while the remaining performers are professional sign language interpreters, ensuring a diverse and authentic representation of sign language expressions within the dataset. The dataset comprises a total of 5,988 continuous sign language videos. According to strict division ratios, there are 4,973 sign language videos in the training set, 515 in the validation set, and 500 in the test set. These videos collectively involve 3,515 Chinese words, covering a broad range of daily communication needs. The video lengths vary widely, ranging from the shortest at 39 frames to the longest at 530 frames, showcasing the richness and flexibility of sign language expression. Detailed statistics of the CE-CNSL dataset are presented in Table \RNum{2}.\par

We also provide a comprehensive listing of key parameters for each performer in Table \RNum{3}, including the recording equipment used, the number of videos, age, and gender, offering detailed baseline information for future research. Given the open setting for background environments, the CE-CNSL dataset includes over 70 different living scenarios, greatly enriching the diversity of the data. To better understand the diversity of backgrounds in the CE-CNSL dataset, we present a series of representative scenes in Figure 2, including indoor, outdoor, park, street, mall, office, and others, all sourced from our dataset. Through these images, one can clearly see the extensive range of scenarios covered by the CE-CNSL dataset, further demonstrating its valuable contribution as a research resource and its significance in advancing the practical application of CSL recognition technology.\par

\begin{table*}[!htbp]
\centering
\caption{Transcription Standards for Converting Chinese Sentences into Sign Language Sentences.}
\label{tab:aStrangeTable5}
\begin{tabular}{c|c}
\hline  
\makecell{Transcription Standards for Converting Chinese Sentences\\ into Sign Language Sentences} & 
Detailed Explanation \\
\hline 
\makecell{When a single word in the dictionary has multiple sign\\ language representations, the specific representation\\ used should be noted at the end of the word.} & 
\makecell{For example: For the word "\begin{CJK*}{UTF8}{gbsn}事情\end{CJK*}" (thing), if there are\\ two sign language representations in the dictionary—"\begin{CJK*}{UTF8}{gbsn}事情1\end{CJK*}"\\ (thing1) and "\begin{CJK*}{UTF8}{gbsn}事情2\end{CJK*}" (thing2)—it is necessary to specify which\\ representation is used during transcription.} \\
\hline 
\makecell{Select the corresponding words based on the actual meaning\\ conveyed by the sentence.} & 
\makecell{For example: The term "\begin{CJK*}{UTF8}{gbsn}不会\end{CJK*}" has two meanings: "do not know" \\ and "impossible." However, if the dictionary only provides a sign\\ for "do not know," during the transcription process, you should\\ choose the signs for "\begin{CJK*}{UTF8}{gbsn}不可能\end{CJK*}(impossible)" or "\begin{CJK*}{UTF8}{gbsn}不行\end{CJK*}(cannot)" based on the\\ actual meaning conveyed by the sentence.} \\
\hline 
\makecell{For terms that involve only the change of numbers, it is\\ assumed that the dictionary contains the term, and no special\\ annotation is required.} & 
\makecell{For example: From January to December, and from Monday to\\ Sunday, the dictionary may only include entries for January and\\ Monday, but if it states "and so on," it is assumed that the\\ dictionary includes representations for all months and days of the\\ week. Therefore, no special annotation is needed for the signs of\\ other months or days.} \\
\hline 
\makecell{For words that do not exist in the dictionary, use regional\\ sign language representations.} & 
\makecell{For example: If the word "\begin{CJK*}{UTF8}{gbsn}英语\end{CJK*}" (English) does not exist\\ in the dictionary, then use the regional sign language\\ representation. Words that use regional sign language\\ representations will be specified in the notes.} \\
\hline 
\makecell{Directional words should be followed by ( ) to specify the\\ subject of expression.} & 
\makecell{For example: "\begin{CJK*}{UTF8}{gbsn}他帮我开门\end{CJK*}" is transcribed as "\begin{CJK*}{UTF8}{gbsn}他/帮(我)/开门.\end{CJK*}"} \\
\hline 
\makecell{For body turns, add [ ] after the action description words\\ to specify the direction.} & 
\makecell{For example: "\begin{CJK*}{UTF8}{gbsn}我想在饮料里放些冰块\end{CJK*}" is transcribed\\ as "\begin{CJK*}{UTF8}{gbsn}喝[左]/冰[右]/块/抓[右]/放[左].\end{CJK*}"} \\
\hline  
\end{tabular}
\end{table*}

\subsection {Dataset annotation}

As a visual language, sign language carries unique grammatical structures and lexical ordering that differ significantly from traditional written or spoken languages, making direct mapping challenging. Therefore, the CE-CNSL dataset introduces deep involvement from professional sign language translators, who are responsible for converting sign glosses into video content. Additionally, the dataset provides spoken language translations as a reference to enhance multidimensional understanding. Notably, we have conducted meticulous word segmentation for the sign glosses, whereas the spoken language translations maintain complete sentence structures without segmentation.\par

The CE-CNSL dataset revolves around topics such as family life, learning, shopping, and daily communication. The sentences are derived from daily communication statements, weather forecasts, news content, and online articles. Regarding the collection process of the dataset, firstly, we collected a large number of written or colloquial sentences from different sources as raw data. Then, we deduplicated the raw data, leaving unique sentences as the final target sentences. Then, 12 professional sign language presenters, including professional sign language translators and hearing-impaired individuals, jointly undertook the task of converting from oral translation to symbol gloss. This process involves a deep understanding of spoken language sentences, followed by their reformation into expressions that conform to sign language grammar, which are then solidified through video demonstrations. All videos and their corresponding annotations are subject to comprehensive review and correction by experienced sign language interpreters after individual tasks are completed, ensuring the accuracy and consistency of the dataset.\par

The current sign language environment in China exists in a state where both standard sign language and regional sign languages coexist. Although efforts have been made to promote a standard sign language dictionary (referred to as the dictionary), a unified standard has yet to be fully established. Against this backdrop, the CE-CNSL dataset adopts an inclusive approach. During the transcription process, in addition to relying on standard sign language, regional sign languages are appropriately integrated, especially when there is no corresponding sign in the standard dictionary for a spoken word. For words that use regional sign languages, we explicitly label them in the annotations, as shown in Table \RNum{4}, to ensure the transparency and traceability of the dataset.\par

The transcription of sign language is not a mechanical process but one guided by a series of norms. For instance, when encountering words that exist in written or spoken language but are missing from the sign language dictionary, we follow the intent of the sentence and select the most appropriate substitute words. When faced with multiple sign language expressions for synonyms listed in the dictionary, we append specific gesture versions after the words to ensure accuracy, as listed in Table \RNum{5}. These guidelines ensure the rigor and practicality of the CE-CNSL dataset, providing researchers with reliable material for their studies.\par

In summary, the CE-CNSL dataset not only demonstrates the richness of sign language videos in quantity but also exhibits professionalism and meticulous annotation and transcription processes, providing a solid data foundation for the advancement of sign language recognition technology. By integrating both standard and regional sign languages, along with rigorous transcription standards, it effectively propels the field of sign language recognition forward.\par

\begin{figure*}
  \begin{center}
  \includegraphics[width=6.0in,height=2.0in]{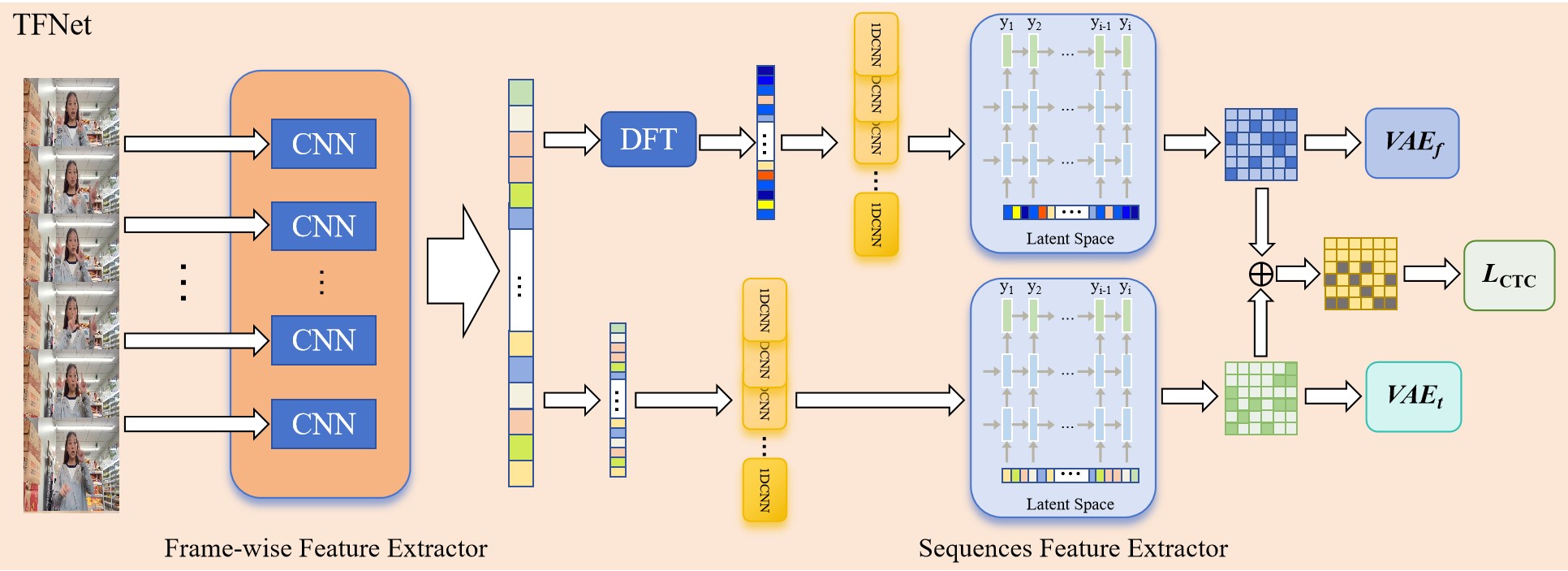}\\
  \caption{Provides an overview of the new TFNet. First, a CNN is used to capture frame-level features, which are then split into two branches. One branch models the temporal sequence using 1DCNN + BiLSTM; the other branch transforms the features into the frequency domain using the Discrete Fourier Transform (DFT) and then models the frequency-domain sequence with 1DCNN + BiLSTM. Finally, a classifier is used to predict the sentence.}\label{fig:ljxy4}
  \end{center}
\end{figure*}

\subsection {Methodology}

This paper proposes a TFNet for CSLR, as illustrated in Figure 3. TFNet primarily consists of two key components: a frame-level feature extraction module and a sequence feature extraction module. Initially, in the frame-level feature extraction stage, we employ a CNN to efficiently extract feature information from frames of sign language videos. Subsequently, in the sequence feature extraction part, the model is designed with two independent yet complementary branches. In one branch, we model the temporal sequence using a combination of 1DCNN and Bidirectional Long Short Term Memory Networks (BiLSTM)\cite{graves2013speech} to capture the temporal dynamics of sign language actions. In the other branch, we transform the temporal features into the frequency domain via Discrete Fourier Transform (DFT)\cite{cai2021frequency} and similarly model the frequency-domain sequence using 1DCNN and BiLSTM to obtain sequence features within the frequency domain. Ultimately, the outputs from these two branches are merged and subjected to a classifier for joint prediction, thereby achieving effective recognition of sign language sentences.\par

\textbf{1) TFNet:} CSLR involves translating continuous sign language videos into written phrases or spoken language understandable by the general population. Let the size of the input continuous sign language video frames be $T$, then the continuous sign language video $V=(x_1,x_2,...,x_T)=\left\{x_t\left|_1^T\in\mathbb{R}^{T\times C\times H\times W}\right.\right\}$, where $x_{t}$ is the image of the t-th frame in the video, $H\times W$ is the dimension of $x_{t}$, and $\text{C}$ is the number of channels. In the TFNet model, we first extract frame-level features and then extract time-frequency sequence features. Let the frame-level feature extractor be $F_{f}$, yielding the feature representation as follows:\par

\begin{equation}
    f_{frame}=F_f(V)\in\mathbb{R}^{T\times C^{\prime}}
\end{equation}

\noindent where $f_{frame}$ is the frame-level feature after extraction, and $C^{\prime}$ is the number of channels after frame-level feature extraction. Then, the frame level feature $f_{frame}$ is input into the sequence feature extractor, which includes two branch structures: one branch focuses on parsing the dynamic characteristics of sign language actions over time based on time-domain feature sequences; Another branch analyzes the dynamic characteristics of sign language actions over time from a frequency domain perspective based on frequency domain feature sequences. This article uses the discrete Fourier transform to transform the eigenvectors into the frequency domain. The following is the basic formula of the discrete Fourier transform:\par

\begin{equation}
    Z[k]=\sum_{n=0}^{N-1}z[n]\cdotp e^{-j2\pi kn/N}
\end{equation}

Among them, $z[n]$ is the input discrete time series, $Z[k]$ is the frequency domain representation of the output, $k=0,1,2,\ldots,N-1$. $j$ is the imaginary unit that satisfies $j^2=-1$. $e^{-j2\pi kn/N}$ is a complex exponential function used to calculate the coefficients of each frequency component. Substitute $f_{frame}$ into formula (2) to calculate the frequency domain sequence feature, and let the discrete Fourier transform be $DFT(\cdot)$, then,\par

\begin{equation}
    f^{'}_{frame}=DFT(f_{frame})\in\mathbb{R}^{T\times C^{\prime}}
\end{equation}

\begin{table*}[!htbp]
\centering
\caption{Performance comparison with other advanced methods on RWTH, RWTH-T, CSL-Daily, and CE-CNSL.}
\label{tab:aStrangeTable6}
\begin{tabular}{c|c|c|c|c|c|c|c|c}
\hline  
\multirow[t]{1}{*}{ Methods } & \multicolumn{2}{c|}{ RWTH } & \multicolumn{2}{c|}{ RWTH-T } & \multicolumn{2}{c|}{ CSL-Daily } & \multicolumn{2}{c}{ CE-CNSL }\\
\cline{2-9}
& Dev(\%) & Test(\%) & Dev(\%) & Test(\%) & Dev(\%) & Test(\%) & Dev(\%) & Test(\%)\\
\hline 
VAC\cite{min2021visual} & 21.2 & 22.3 & - & - & - & - & 45.1 & 43.3\\
\hline 
MSTNet\cite{zhu2024multiscale} & 20.3 & 21.4 & - & - & - & - & 54.4 & 53.0\\
\hline 
SEN\cite{hu2023self} & 19.5 & 21.0 & 19.3 & 20.7 & 31.1 & 30.7 & 46.5 & 45.3\\
\hline 
AdaSize\cite{hu2024scalable} & 19.7 & 20.9 & 19.7 & 21.2 & 31.3 & 30.9 & - & -\\
\hline 
STMC\cite{zhou2020spatial} & 21.1 & 20.7 & 19.6 & 21.0 & - & - \\
\hline 
STENet\cite{yin2023spatial} & 19.3 & 20.3 & 19.4 & 21.1 & 28.9 & 28.9 & - & -\\
\hline 
HST-GNN\cite{kan2022sign} & 19.5 & 19.8 & 20.1 & 20.3 & - & - & - & -\\
\hline 
DSE\cite{wang2024dynamical} & 18.6 & 19.8 & 18.9 & 19.9 & 28.6 & 27.6 & - & -\\
\hline 
TB-Net\cite{liu2024tb} & 18.9 & 19.6 & 18.8 & 20.0 & 28.4 & 28.2 & - & -\\
\hline 
CorrNet\cite{hu2023continuous} & 18.8 & 19.4 & 18.9 & 20.5 & 30.6 & 30.1 & 47.2 & 46.5\\
\hline 
TCNet\cite{lu2024tcnet} & 18.1 & 18.9 & 18.3 & 19.4 & 29.7 & 29.3 & - & -\\
\hline 
MAM-FSD\cite{zhu2024continuous} & 19.2 & 18.8 & 18.2 & 19.4 & 25.8 & 24.5 & 44.9 & 44.7\\
\hline 
TwoStream-SLR\cite{chen2022two} & 18.4 & 18.8 & 17.7 & 19.3 & 25.4 & 25.3 & - & -\\
\hline 
TFNet & 18.7 & 18.6 & 18.0 & 19.1 & 25.1 & 23.5 & 42.1 & 41.9\\
\hline  
\end{tabular}
\end{table*}

\noindent where $f^{'}_{frame}$ is the result of the discrete Fourier transform of the temporal feature sequence.\par

One branch directly inputs $f_{frame}$ into the time-domain sequence feature extractor, which is composed of a hybrid neural network consisting of 1DCNN and BiLSTM. Firstly, $f_{frame}$ is transposed and passed through 1DCNN, where each convolution is followed by a BN layer, ReLU activation function, and max pooling layer.\par

\begin{equation}    
\mathrm{g}_i=\mathrm{MaxPool}(\mathrm{ReLU}(\mathrm{BN}(\sum_{k=1}^K\left(f_{i+k-1}\cdot w_k\right)+b)))
\end{equation}

Among them, $w$ is the weight of the convolution kernel, $K$ is the number of convolution kernels, $b$ is the bias term, $g_i$ is the i-th output data, $f_i$ is the i-th data of the $f_{frame}$ feature vector, and $i\in[0,T]$. All output data are defined as $f_{gloss}\in\mathbb{R}^{C^{\prime\prime}\times T^{\prime}}$, where $T^{\prime}$ is the length of the downsampled sequence features in the time dimension, and $C^{\prime\prime}$ is the size of the number of channels extracted from the temporal features. Then transpose it and output it to BiLSTM.\par

\begin{equation}
f_{temporal}=BiLSTM\left(f_{gloss}\right)\in\mathbb{R}^{T^{\prime}\times C^{\prime\prime}}
\end{equation}

Among them, $f_{temporal}$ is the time-domain sequence feature extracted from the sequence feature. Another branch inputs $f^{'}_{frame}$ into the frequency domain sequence feature extractor $F_{\mathrm{sequences}_f}$, and the methods used by these two sequence feature extractors are the same (both 1DCNN+BiLSTM), which yields,\par

\begin{equation}
    f_{freq}=F_{\mathrm{sequences}_f}(f^{'}_{frame})\in\mathbb{R}^{T^{\prime}\times C^{\prime\prime}}
\end{equation}

Among them, $f_{freq}$ is the extracted frequency domain sequence feature, and the size of the frequency domain sequence feature is consistent with that of the time domain sequence feature. Finally, the time-domain sequence feature $f_{temporal}$ and frequency-domain sequence feature $f_{freq}$ are summed and identified through a fully connected layer.\par

\begin{equation}
    f_{calssification}=F_{linear}(f_{temporal}+f_{freq})\in\mathbb{R}^{T^{\prime}\times l}
\end{equation}

\noindent Where $f_{classification}$  is the final recognition result, and $l$ is the number of words.\par

\textbf{2) Loss Function:} The overall loss function $L_{sum}$ of the TFNet model consists of three parts: two auxiliary VAE loss functions , $L_{VAE_t}$ and $L_{VAE_f}$, and a final CTC (connectionist temporary classification)\cite{graves2006connectionist} loss function. The VAE loss function was proposed by Min et al.\cite{min2021visual} in 2021, which includes two parts: visual enhancement and visual alignment loss.\par

(1) Visual Enhancement Loss\par
In order to enhance the feature extraction capability of the feature extractor, researchers have implemented it by providing an auxiliary loss to the feature extractor, which is the Visual Enhancement (VE) loss. This auxiliary loss is designed to allow the feature extractor to make predictions based on local visual information, which significantly improves the model's ability to capture details compared to traditional methods. Specifically, a CTC loss is introduced as a VE loss on the auxiliary classifier in the feature extraction stage with the formula:\par

\begin{equation}
    L_{VE}=L_{CTC}=-\log p(l\mid X;\theta)
\end{equation}

\noindent where $l$ denotes the label, $X$ denotes the input sequence, and $\theta$ denotes the model parameters.

(2) Visual alignment loss\par
Since the VE loss function lacks contextual information and is independent of the main loss, this can lead to misalignment between the two classifiers, which affects the overall performance of the model. To circumvent this problem, the researchers used a Visual Alignment (VA) loss function for optimization.The design of the VA loss function is based on the idea of knowledge distillation, which treats the whole network as a teacher model and the visual feature extractors as student models. In this way, the student model can learn more contextual information and deeper feature representations from the teacher model, thus improving the accuracy and robustness of its prediction. And the high temperature parameter $\tau$ is introduced to soften the probability distribution of the peak response when calculating the VA loss. This softening operation can make the probability distribution smoother, thus reducing the risk of overfitting.\par

\begin{equation}
    L_{VA}=KL\left(\mathrm{softmax}(\frac{Z}{\tau}),\mathrm{softmax}(\frac{\tilde{Z}}{\tau})\right)
\end{equation}

\noindent where $Z$ denotes the predicted value and $\tilde{Z}$ denotes the standard output.\par

Specifically, $L_{VAE_{t}}$ is dedicated to applying the VAE loss function to the time-domain sequence features, which aims to achieve a sharp capture and efficient representation of the temporal dynamics; meanwhile, $L_{VAE_f}$ focuses on executing the VAE loss function to the frequency-domain sequence features, which aims to strengthen the model's effectiveness in processing the frequency-domain information. The final CTC loss function, which is used to calculate the difference between the predicted sequences and the real labels, guides the model to learn the correct sign language recognition path, then $L_{sum}$ is defined as:\par

\begin{equation}
    L_{sum}=L_{VAE_t}+L_{VAE_f}+L_{CTC}
\end{equation}

\begin{table*}[!htbp]
\centering
\caption{Qualitative results for the CE-CNSL dataset, with red font indicating identification errors.}
\label{tab:aStrangeTable7}
\begin{tabular}{c|c|c|c|c}
\hline  
CSLR & Test-Case-1 & WER(\%) & Test-Case 2 & WER(\%) \\
\hline 
Original sentence & \begin{CJK*}{UTF8}{gbsn}他从小就相当律师。\end{CJK*} & - & \begin{CJK*}{UTF8}{gbsn}我可以支持你去运动。\end{CJK*} & - \\
\hline 
Gloss ground truth & \begin{CJK*}{UTF8}{gbsn}他/小孩时间/开始/做/律师/希望/。\end{CJK*} & - & \begin{CJK*}{UTF8}{gbsn}我/可以支持/你/去/运动/。\end{CJK*} & - \\
\hline 
Pred. (SEN) & \begin{CJK*}{UTF8}{gbsn}他/时间/开始/。\end{CJK*} & 50.0 & \begin{CJK*}{UTF8}{gbsn}我/可以你/\textcolor{red}{经济}/\textcolor{red}{木头}/。\end{CJK*} & 42.9 \\
\hline
Pred. (CorrNet) & \begin{CJK*}{UTF8}{gbsn}他时间/开始/做/希望/。\end{CJK*} & 25.0 & \begin{CJK*}{UTF8}{gbsn}我/可以/你/。\end{CJK*} & 42.9 \\
\hline 
Pred. (VAC) & \begin{CJK*}{UTF8}{gbsn}他/小孩时间/开始/\textcolor{red}{工作}/\textcolor{red}{困难}/。\end{CJK*} & 37.5 & \begin{CJK*}{UTF8}{gbsn}我/可以/你/\textcolor{red}{好}/\textcolor{red}{锻炼}/。\end{CJK*} & 42.9 \\
\hline 
Pred. (MAM-FSD) & \begin{CJK*}{UTF8}{gbsn}他/小孩时间/开始/做/律师/希望/。\end{CJK*} & 0.0 & \begin{CJK*}{UTF8}{gbsn}我/可以/支持/你/\textcolor{red}{锻炼}/。\end{CJK*} & 28.6 \\
\hline 
Pred (TFNet) & \begin{CJK*}{UTF8}{gbsn}他/小孩时间/开始/做/律师/希望/。\end{CJK*} & 0.0 & \begin{CJK*}{UTF8}{gbsn}我/可以/支持/你/去/运动/。\end{CJK*} & 0.0 \\
\hline  
\end{tabular}
\end{table*}

\begin{table*}[!htbp]
\centering
\caption{Ablation study of different domain sequence features.}
\label{tab:aStrangeTable8}
\begin{tabular}{c|c|c}
\hline  
Domain type & Dev(\%) & Test(\%) \\
\hline 
Temporal domain sequence features & 19.5 & 19.1 \\
\hline 
Frequency domain sequence features & 19.7 & 20.0 \\
\hline 
Time-frequency domain sequence features & 18.7 & 18.6 \\
\hline  
\end{tabular}
\end{table*}

\begin{table*}[!htbp]
\centering
\caption{Ablation study of different sequence feature extractors (original sequence feature extractors with different models retrained).}
\label{tab:aStrangeTable9}
\begin{tabular}{c|c|c|c|c}
\hline  
\multirow[t]{1}{*}{ Model } & \multicolumn{2}{c|}{ Original sequence feature extractor } & \multicolumn{2}{c}{ Time-frequency sequence feature extractor } \\
\cline{2-5}
& Dev(\%) & Test(\%) & Dev(\%) & Test(\%) \\
\hline 
VAC & 21.2 & 22.3 & 19.9 & 20.2 \\
\hline 
MSTNet & 20.3 & 21.4 & 19.7 & 19.8 \\
\hline 
SEN & 19.5 & 21.0 & 19.4 & 19.3 \\
\hline 
CorrNet & 18.8 & 19.4 & 19.2 & 19.1 \\
\hline 
MAM-FSD & 19.2 & 18.8 & 18.7 & 18.6 \\
\hline  
\end{tabular}
\end{table*}

\begin{table*}[!htbp]
\centering
\caption{Ablation study of different loss combinations.}
\label{tab:aStrangeTable10}
\begin{tabular}{c|c|c|c|c|c}
\hline  
Number & $loss_{CTC}$ & $loss_{VAE_t}$ & $loss_{VAE_f}$ & Dev(\%) & Test(\%) \\
\hline 
1 & $\checkmark$ & - & - & 21.3 & 21.4 \\
\hline 
2 & $\checkmark$ & $\checkmark$ & - & 19.3 & 19.0 \\
\hline 
3 & $\checkmark$ & - & $\checkmark$ & 19.0 & 19.3 \\
\hline 
4 & $\checkmark$ & $\checkmark$ & $\checkmark$ & 18.7 & 18.6 \\
\hline  
\end{tabular}
\end{table*}
\section{Experiment}

\subsection {Dataset and judgment criteria}

In addition to the CE-CNSL dataset constructed in this paper, we conducted experiments on three other large-scale publicly available datasets, including RWTH\cite{koller2015continuous}, RWTH-T\cite{camgoz2018neural}, and CSL-Daily\cite{zhou2021improving}, aimed at verifying the wide applicability and effectiveness of the proposed method. The RWTH dataset contains a total of 6841 different video clips, which is derived from sign language videos recorded by the German Weather Radio and Television (DWRT). The official pre-defined splits consist of a training set (5,672 videos), a validation set (540 videos), and a test set (629 videos). The RWTH-T dataset is an extended version of the RWTH dataset with an expanded size, where the dataset is subdivided into 7,096 training videos, 519 validation videos, and 642 test videos. The CSL-Daily dataset is a large Chinese sign language dataset with an annotated vocabulary of 2,000 words and a Chinese text vocabulary of 2,343 words. The dataset consists of 18,401 training samples, 1,077 validation samples, and 1,176 test samples.\par

We use Word Error Rate (WER)\cite{koller2015continuous} as the evaluation criterion, which is widely applied in CSLR. WER is defined as the sum of the minimum number of insertion, substitution, and deletion operations required to transform the recognized sequence into the standard reference sequence. A lower WER indicates a better recognition performance, which is defined as follows.\par

\begin{equation}
    WER=100\%\times\frac{ins+del+sub}{sum}
\end{equation}

\noindent where $ins$ represents the number of words to be inserted, $del$ represents the number of words to be deleted, $sub$ represents the number of words to be replaced, and $sum$ represents the total number of words in the label.

\subsection {Implementation rules}

In our experimental implementation, the frame-level feature extraction component of the TFNet model utilizes the backbone of the MAM-FSD model\cite{zhu2024continuous}. We train the entire model using the Adam optimizer\cite{kingma2014adam}, with an initial learning rate and weight factor set to $10^{-4}$, and a batch size of 2. The graphics card used in the experiments is the RTX3090Ti, equipped with 24GB of GPU-dedicated memory. Data augmentation is performed using random cropping and random flipping. For random cropping, the input data size is $256\times256$, and the cropped size after random cropping is $224\times224$. For random flipping, the flipping probability is set to 0.5. Flipping and cropping are applied to the video sequence. Additionally, temporal enhancement is conducted by randomly lengthening or shortening the length of the video sequence by ±20\%. Training is carried out for a total of 55 epochs, with the learning rate reduced by 80\% at the 35th and 45th epochs. During model testing, only central cropping is used for data augmentation and a beam search algorithm with a beam width of 10 is employed in the final CTC decoding phase.\par

\subsection {Experimental results}

The TFNet model is tested on RWTH, RWTH-T, CSL-Daily, and CE-CNSL datasets, with the model recognition accuracies presented in Tables \RNum{6} and \RNum{7}.\par

As can be seen from Table \RNum{6}, TFNet achieves highly competitive results compared to other advanced models. On the RWTH dataset, TFNet attains a validation error rate of 18.7\% and a test error rate of 18.6\%, performing excellently among all compared methods. Compared to the closest method, TwoStream-SLR\cite{chen2022two}, TFNet achieves a lower error rate on the test set, indicating that TFNet not only fits the training data well but also generalizes effectively to unseen data. On the RWTH-T dataset, TFNet has a validation error rate of 18.0\% and a test error rate of 19.1\%, showing equally good performance. Relative to TwoStream-SLR, TFNet exhibits lower errors on the validation set, suggesting that TFNet is more reasonable in terms of model selection and hyperparameter tuning. Additionally, TFNet performs slightly better than TwoStream-SLR on the test set, which further confirms the generalization ability of TFNet. On the CSL-Daily dataset, TFNet achieves the best performance on this dataset compared to other methods with a validation error rate of 25.1\% and a test error rate of 23.5\%. Compared with the closest method TwoStream-SLR, TFNet shows better performance on the validation set and a significant improvement on the test set, indicating that TFNet possesses strong adaptability and robustness when dealing with large-scale complex sign language data.\par

In Table \RNum{6}, we compare the performance of various CSLR methods on the CE-CNSL dataset. As shown in Table \RNum{6}, the test results of our proposed model on CE-CNSL indicate its superior performance. We also analyze the performance of TFNet on CE-CNSL from both quantitative and qualitative perspectives.\par

\textbf{Quantitative Analysis:} To evaluate the effectiveness of the proposed model, we conduct comparative experiments on the CE-CNSL dataset. The results reveal significant advantages of our method. According to the data in Table \RNum{6}, our model achieve WER of 42.1\% and 41.9\% on the validation and test sets respectively. Compared to the closest CSLR model MAM-FSD, which have WER values of 44.9\% and 44.7\% on the validation and test sets, our model demonstrates a reduction of 2.8\%, highlighting the superiority of our proposed model in sign language recognition.\par

\textbf{Qualitative Analysis:} To delve deeper and provide an intuitive demonstration of the performance of different models in recognizing Chinese continuous sign language datasets in complex environments, we select several CSLR models and analyze their test results on the CE-CNSL dataset. The analyzed results in Table \RNum{7}, aiming to offer a more intuitive perspective. With the results in Table \RNum{7}, we find that the TFNet model has the lowest recognition error rate, marking it as having the best recognition performance. This result reflects the model's excellent capability in handling complex backgrounds and continuous sign language videos.\par

\subsection {Ablation experiment}

In this section, we conduct ablation experiments on the RWTH dataset to further validate the effectiveness of the model components. In the ablation experiments, we use WER as the metric, with smaller WER indicating better performance.\par

\textbf{Ablation of different domain sequence features.} We conduct experiments by combining temporal, spectral, and time-frequency domain sequence features after frame-level feature extraction to investigate the role of different domain sequence feature combinations in CSLR. The experimental results are shown in Table \RNum{8}. As can be observed from Table \RNum{8}, when using only temporal domain sequence features for recognition, the WERs for the validation and test sets are 19.5\% and 19.1\%, respectively. When using only spectral domain sequence features for recognition, the WERs are 19.7\% and 20.0\%, respectively. This indicates that using only temporal domain sequence features yields slightly better performance than using only spectral domain sequence features, with WERs reduced by 0.2\% and 0.9\%, respectively. Finally, by integrating temporal and spectral domain sequence features to form time-frequency domain sequence features for recognition, the results showed further improvements. Compared to using only temporal domain sequence features, the WER decreased by 0.8\% and 0.5\%, respectively, achieving even better results. This suggests that the integration of time-frequency domain sequence features can effectively enhance the performance of CSLR systems.\par

\textbf{Ablation of different sequence feature extractors.} Under the condition that the frame-level feature extractors of different models remain unchanged, we investigate the effect of replacing their respective sequence feature extractors with time-frequency sequence feature extractors. The experimental results are shown in Table \RNum{9}. From Table \RNum{9}, it can be seen that for different models, recognition using time-frequency sequence features results in a decrease in WER on the test set compared to the original sequence features. Specifically, the WER of MAM-FSD is reduced by 0.2\%, the WER of CorrNet is reduced by 0.3\%, and the WER of SEN is reduced by 1.7\%. On the validation set, only the BER of the CorrNet model increases by 0.4\%, while the performance of the remaining models on the validation set improves with the use of the time-frequency sequence feature extractor. This experimental result proves the sophistication of the time-frequency sequence feature extractor we used.\par

\textbf{Ablation of Different Loss Combinations.} We study the effects of various loss function combinations in the TFNet model by experimenting with different combinations of the CTC loss function $L_{CTC}$, the temporal domain sequence feature VAE loss function $L_{VAE_{t}}$ and the spectral domain sequence feature VAE loss function $L_{VAE_f}$. The experimental results are shown in Table \RNum{10}. As can be observed from Table \RNum{10}, using only $L_{CTC}$ results in WERs of 21.3\% and 21.4\% on the validation and test sets, respectively. When $L_{CTC}$ is combined with either of the VAE loss functions, there is an improvement in performance. Combining $L_{CTC}$ with $L_{VAE_{t}}$, the WERs on the validation and test sets decrease by 2.0\% and 2.4\%, respectively. Combining $L_{CTC}$ with $L_{VAE_f}$, the WERs on the validation and test sets decrease by 2.3\% and 2.1\%, respectively. When all three loss functions are combined, the WER reaches its minimum on both the validation and test sets, at 18.7\% and 18.6\%, respectively. The result indicates that combining the CTC loss function with the time-frequency domain sequence feature VAE loss functions can significantly enhance the performance of CSLR systems.\par

\section{Conclusion}

In this paper, we introduce a novel CE-CNSL dataset designed for complex environments, aiming to facilitate the transition of CSLR from laboratory settings to real-life scenarios. The CE-CNSL encompasses a rich variety of daily communication contexts, comprising a total of 5,988 continuous sign language videos, meticulously divided into a training set of 4,973 sign language videos, a validation set of 515 sign language videos, and a test set of 500 sign language videos. The dataset covers 3,515 Chinese words with a variety of video backgrounds, introducing more than 70 different environments, providing unprecedented complexity and realism for model training. Considering the impact of complex backgrounds on CSLR, we also propose the TFNet to achieve efficient and accurate recognition. On the validation and test sets of the CE-CNSL, the TFNet achieves WERs of 42.1\% and 41.9\%, respectively. Experimental results indicate that the TFNet significantly improves recognition accuracy. Additionally, we applied our proposed method to three publicly available continuous sign language datasets and achieved highly competitive results.\par 

\section*{Acknowledgment}

This work was supported in part by the Development Project of Ship Situational Intelligent Awareness System, China under Grant MC-201920-X01, in part by the National Natural Science Foundation of China under Grant 61673129. \par

~\\\par
\textbf{Data availability} The datasets used in the paper are cited properly.\par

~\\\par
\textbf{Ethics approval} The project uses publicly available data that has been subject to an exemption.\par

\section*{Declarations}

\textbf{Conflict of interest} The authors declare that they have no known competing financial interests or personal relationships that could have appeared to influence the work reported in this paper.\par


%





\ifCLASSOPTIONcaptionsoff
  \newpage
\fi





\bibliographystyle{IEEEtran}
\bibliography{IEEEabrv,Bibliography}

@IEEEtranBSTCTL{IEEEexample:BSTcontrol,
    CTLuse_article_number = "yes",
    CTLuse_paper = "yes",
    CTLuse_forced_etal = "no",
    CTLmax_names_forced_etal = "50",
    CTLnames_show_etal = "50",
    CTLuse_alt_spacing = "yes",
    CTLalt_stretch_factor = "4",
    CTLdash_repeated_names = "yes",
    CTLname_format_string = "{f.~}{vv~}{ll}{, jj}",
    CTLname_latex_cmd = "",
    CTLname_url_prefix = "[Online]. Available:"
 }

@inproceedings{pu2020boosting,
  title={Boosting continuous sign language recognition via cross modality augmentation},
  author={Pu, Junfu and Zhou, Wengang and Hu, Hezhen and Li, Houqiang},
  booktitle={Proceedings of the 28th ACM international conference on multimedia},
  pages={1497--1505},
  year={2020}
}

@article{wadhawan2021sign,
  title={Sign language recognition systems: A decade systematic literature review},
  author={Wadhawan, Ankita and Kumar, Parteek},
  journal={Archives of computational methods in engineering},
  volume={28},
  pages={785--813},
  year={2021},
  publisher={Springer}
}

@inproceedings{cui2017recurrent,
  title={Recurrent convolutional neural networks for continuous sign language recognition by staged optimization},
  author={Cui, Runpeng and Liu, Hu and Zhang, Changshui},
  booktitle={Proceedings of the IEEE conference on computer vision and pattern recognition},
  pages={7361--7369},
  year={2017}
}

@article{uthus2024youtube,
  title={Youtube-asl: A large-scale, open-domain american sign language-english parallel corpus},
  author={Uthus, Dave and Tanzer, Garrett and Georg, Manfred},
  journal={Advances in Neural Information Processing Systems},
  volume={36},
  year={2024}
}

@inproceedings{duarte2021how2sign,
  title={How2sign: a large-scale multimodal dataset for continuous american sign language},
  author={Duarte, Amanda and Palaskar, Shruti and Ventura, Lucas and Ghadiyaram, Deepti and DeHaan, Kenneth and Metze, Florian and Torres, Jordi and Giro-i-Nieto, Xavier},
  booktitle={Proceedings of the IEEE/CVF conference on computer vision and pattern recognition},
  pages={2735--2744},
  year={2021}
}

@article{ozdemir2020bosphorussign22k,
  title={Bosphorussign22k sign language recognition dataset},
  author={{\"O}zdemir, O{\u{g}}ulcan and K{\i}nd{\i}ro{\u{g}}lu, Ahmet Alp and Camg{\"o}z, Necati Cihan and Akarun, Lale},
  journal={arXiv preprint arXiv:2004.01283},
  year={2020}
}

@article{wang2014similarity,
  title={Similarity assessment model for Chinese sign language videos},
  author={Wang, Li-Chun and Wang, Ru and Kong, De-Hui and Yin, Bao-Cai},
  journal={IEEE Transactions on Multimedia},
  volume={16},
  number={3},
  pages={751--761},
  year={2014},
  publisher={IEEE}
}

@article{koller2015continuous,
  title={Continuous sign language recognition: Towards large vocabulary statistical recognition systems handling multiple signers},
  author={Koller, Oscar and Forster, Jens and Ney, Hermann},
  journal={Computer Vision and Image Understanding},
  volume={141},
  pages={108--125},
  year={2015},
  publisher={Elsevier}
}

@inproceedings{camgoz2018neural,
  title={Neural sign language translation},
  author={Camgoz, Necati Cihan and Hadfield, Simon and Koller, Oscar and Ney, Hermann and Bowden, Richard},
  booktitle={Proceedings of the IEEE conference on computer vision and pattern recognition},
  pages={7784--7793},
  year={2018}
}

@article{adaloglou2021comprehensive,
  title={A comprehensive study on deep learning-based methods for sign language recognition},
  author={Adaloglou, Nikolas and Chatzis, Theocharis and Papastratis, Ilias and Stergioulas, Andreas and Papadopoulos, Georgios Th and Zacharopoulou, Vassia and Xydopoulos, George J and Atzakas, Klimnis and Papazachariou, Dimitris and Daras, Petros},
  journal={IEEE transactions on multimedia},
  volume={24},
  pages={1750--1762},
  year={2021},
  publisher={IEEE}
}

@inproceedings{zhou2021improving,
  title={Improving sign language translation with monolingual data by sign back-translation},
  author={Zhou, Hao and Zhou, Wengang and Qi, Weizhen and Pu, Junfu and Li, Houqiang},
  booktitle={Proceedings of the IEEE/CVF Conference on Computer Vision and Pattern Recognition},
  pages={1316--1325},
  year={2021}
}

@article{xie2023multi,
  title={Multi-scale local-temporal similarity fusion for continuous sign language recognition},
  author={Xie, Pan and Cui, Zhi and Du, Yao and Zhao, Mengyi and Cui, Jianwei and Wang, Bin and Hu, Xiaohui},
  journal={Pattern Recognition},
  volume={136},
  pages={109233},
  year={2023},
  publisher={Elsevier}
}

@article{wei2020semantic,
  title={Semantic boundary detection with reinforcement learning for continuous sign language recognition},
  author={Wei, Chengcheng and Zhao, Jian and Zhou, Wengang and Li, Houqiang},
  journal={IEEE Transactions on Circuits and Systems for Video Technology},
  volume={31},
  number={3},
  pages={1138--1149},
  year={2020},
  publisher={IEEE}
}

@article{aditya2022novel,
  title={Novel spatio-temporal continuous sign language recognition using an attentive multi-feature network},
  author={Aditya, Wisnu and Shih, Timothy K and Thaipisutikul, Tipajin and Fitriajie, Arda Satata and Gochoo, Munkhjargal and Utaminingrum, Fitri and Lin, Chih-Yang},
  journal={Sensors},
  volume={22},
  number={17},
  pages={6452},
  year={2022},
  publisher={MDPI}
}

@inproceedings{feng2022sign,
  title={Sign language translation based on new continuous sign language dataset},
  author={Feng, Shi and Yuan, Tiantian},
  booktitle={2022 IEEE International Conference on Artificial Intelligence and Computer Applications (ICAICA)},
  pages={491--494},
  year={2022},
  organization={IEEE}
}

@inproceedings{von2010signum,
  title={Signum database: Video corpus for signer-independent continuous sign language recognition},
  author={von Agris, Ulrich and Kraiss, Karl-Friedrich},
  booktitle={4th Workshop on the Representation and Processing of Sign Languages: Corpora and Sign Language Technologies},
  pages={243--246},
  year={2010}
}

@inproceedings{huang2018video,
  title={Video-based sign language recognition without temporal segmentation},
  author={Huang, Jie and Zhou, Wengang and Zhang, Qilin and Li, Houqiang and Li, Weiping},
  booktitle={Proceedings of the AAAI Conference on Artificial Intelligence},
  volume={32},
  number={1},
  year={2018}
}

@article{niu2024hong,
  title={A Hong Kong Sign Language Corpus Collected from Sign-interpreted TV News},
  author={Niu, Zhe and Zuo, Ronglai and Mak, Brian and Wei, Fangyun},
  journal={arXiv preprint arXiv:2405.00980},
  year={2024}
}

@article{rastgoo2021sign,
  title={Sign language recognition: A deep survey},
  author={Rastgoo, Razieh and Kiani, Kourosh and Escalera, Sergio},
  journal={Expert Systems with Applications},
  volume={164},
  pages={113794},
  year={2021},
  publisher={Elsevier}
}

@inproceedings{cheng2020fully,
  title={Fully convolutional networks for continuous sign language recognition},
  author={Cheng, Ka Leong and Yang, Zhaoyang and Chen, Qifeng and Tai, Yu-Wing},
  booktitle={Computer Vision--ECCV 2020: 16th European Conference, Glasgow, UK, August 23--28, 2020, Proceedings, Part XXIV 16},
  pages={697--714},
  year={2020},
  organization={Springer}
}

@inproceedings{hu2022temporal,
  title={Temporal lift pooling for continuous sign language recognition},
  author={Hu, Lianyu and Gao, Liqing and Liu, Zekang and Feng, Wei},
  booktitle={European conference on computer vision},
  pages={511--527},
  year={2022},
  organization={Springer}
}

@inproceedings{hu2023continuous,
  title={Continuous sign language recognition with correlation network},
  author={Hu, Lianyu and Gao, Liqing and Liu, Zekang and Feng, Wei},
  booktitle={Proceedings of the IEEE/CVF Conference on Computer Vision and Pattern Recognition},
  pages={2529--2539},
  year={2023}
}

@article{zhu2024continuous,
  title={Continuous Sign Language Recognition Based on Motor attention mechanism and frame-level Self-distillation},
  author={Zhu, Qidan and Li, Jing and Yuan, Fei and Gan, Quan},
  journal={arXiv preprint arXiv:2402.19118},
  year={2024}
}

@article{zhu2024multiscale,
  title={Multiscale temporal network for continuous sign language recognition},
  author={Zhu, Qidan and Li, Jing and Yuan, Fei and Gan, Quan},
  journal={Journal of Electronic Imaging},
  volume={33},
  number={2},
  pages={023059--023059},
  year={2024},
  publisher={Society of Photo-Optical Instrumentation Engineers}
}

@inproceedings{zhou2020spatial,
  title={Spatial-temporal multi-cue network for continuous sign language recognition},
  author={Zhou, Hao and Zhou, Wengang and Zhou, Yun and Li, Houqiang},
  booktitle={Proceedings of the AAAI conference on artificial intelligence},
  volume={34},
  number={07},
  pages={13009--13016},
  year={2020}
}

@inproceedings{yin2023gloss,
  title={Gloss attention for gloss-free sign language translation},
  author={Yin, Aoxiong and Zhong, Tianyun and Tang, Li and Jin, Weike and Jin, Tao and Zhao, Zhou},
  booktitle={Proceedings of the IEEE/CVF conference on computer vision and pattern recognition},
  pages={2551--2562},
  year={2023}
}

@inproceedings{koller2017re,
  title={Re-sign: Re-aligned end-to-end sequence modelling with deep recurrent CNN-HMMs},
  author={Koller, Oscar and Zargaran, Sepehr and Ney, Hermann},
  booktitle={Proceedings of the IEEE conference on computer vision and pattern recognition},
  pages={4297--4305},
  year={2017}
}

@article{hu2023signbert+,
  title={Signbert+: Hand-model-aware self-supervised pre-training for sign language understanding},
  author={Hu, Hezhen and Zhao, Weichao and Zhou, Wengang and Li, Houqiang},
  journal={IEEE Transactions on Pattern Analysis and Machine Intelligence},
  volume={45},
  number={9},
  pages={11221--11239},
  year={2023},
  publisher={IEEE}
}

@inproceedings{wei2023improving,
  title={Improving continuous sign language recognition with cross-lingual signs},
  author={Wei, Fangyun and Chen, Yutong},
  booktitle={Proceedings of the IEEE/CVF International Conference on Computer Vision},
  pages={23612--23621},
  year={2023}
}

@inproceedings{graves2013speech,
  title={Speech recognition with deep recurrent neural networks},
  author={Graves, Alex and Mohamed, Abdel-rahman and Hinton, Geoffrey},
  booktitle={2013 IEEE international conference on acoustics, speech and signal processing},
  pages={6645--6649},
  year={2013},
  organization={Ieee}
}

@inproceedings{cai2021frequency,
  title={Frequency domain image translation: More photo-realistic, better identity-preserving},
  author={Cai, Mu and Zhang, Hong and Huang, Huijuan and Geng, Qichuan and Li, Yixuan and Huang, Gao},
  booktitle={Proceedings of the IEEE/CVF International Conference on Computer Vision},
  pages={13930--13940},
  year={2021}
}

@inproceedings{graves2006connectionist,
  title={Connectionist temporal classification: labelling unsegmented sequence data with recurrent neural networks},
  author={Graves, Alex and Fern{\'a}ndez, Santiago and Gomez, Faustino and Schmidhuber, J{\"u}rgen},
  booktitle={Proceedings of the 23rd international conference on Machine learning},
  pages={369--376},
  year={2006}
}

@inproceedings{min2021visual,
  title={Visual alignment constraint for continuous sign language recognition},
  author={Min, Yuecong and Hao, Aiming and Chai, Xiujuan and Chen, Xilin},
  booktitle={Proceedings of the IEEE/CVF international conference on computer vision},
  pages={11542--11551},
  year={2021}
}

@inproceedings{hu2023self,
  title={Self-emphasizing network for continuous sign language recognition},
  author={Hu, Lianyu and Gao, Liqing and Liu, Zekang and Feng, Wei},
  booktitle={Proceedings of the AAAI Conference on Artificial Intelligence},
  volume={37},
  number={1},
  pages={854--862},
  year={2023}
}

@article{hu2024scalable,
  title={Scalable frame resolution for efficient continuous sign language recognition},
  author={Hu, Lianyu and Gao, Liqing and Liu, Zekang and Feng, Wei},
  journal={Pattern Recognition},
  volume={145},
  pages={109903},
  year={2024},
  publisher={Elsevier}
}

@article{yin2023spatial,
  title={Spatial temporal enhanced network for continuous sign language recognition},
  author={Yin, Wenjie and Hou, Yonghong and Guo, Zihui and Liu, Kailin},
  journal={IEEE Transactions on Circuits and Systems for Video Technology},
  year={2023},
  publisher={IEEE}
}

@inproceedings{kan2022sign,
  title={Sign language translation with hierarchical spatio-temporal graph neural network},
  author={Kan, Jichao and Hu, Kun and Hagenbuchner, Markus and Tsoi, Ah Chung and Bennamoun, Mohammed and Wang, Zhiyong},
  booktitle={Proceedings of the IEEE/CVF winter conference on applications of computer vision},
  pages={3367--3376},
  year={2022}
}

@article{wang2024dynamical,
  title={Dynamical Semantic Enhancement Network for Continuous Sign Language Recognition},
  author={WANG, SUYANG and GUO, LEMING and XUE, WANLI},
  year={2024}
}

@article{liu2024tb,
  title={TB-Net: Intra-and inter-video correlation learning for continuous sign language recognition},
  author={Liu, Jingze and Xue, Wanli and Zhang, Kaihua and Yuan, Tiantian and Chen, Shengyong},
  journal={Information Fusion},
  volume={109},
  pages={102438},
  year={2024},
  publisher={Elsevier}
}

@inproceedings{lu2024tcnet,
  title={TCNet: Continuous Sign Language Recognition from Trajectories and Correlated Regions},
  author={Lu, Hui and Salah, Albert Ali and Poppe, Ronald},
  booktitle={Proceedings of the AAAI Conference on Artificial Intelligence},
  volume={38},
  number={4},
  pages={3891--3899},
  year={2024}
}

@article{chen2022two,
  title={Two-stream network for sign language recognition and translation},
  author={Chen, Yutong and Zuo, Ronglai and Wei, Fangyun and Wu, Yu and Liu, Shujie and Mak, Brian},
  journal={Advances in Neural Information Processing Systems},
  volume={35},
  pages={17043--17056},
  year={2022}
}

@article{kingma2014adam,
  title={Adam: A method for stochastic optimization},
  author={Kingma, Diederik P},
  journal={arXiv preprint arXiv:1412.6980},
  year={2014}
}

\vfill


\end{document}